\documentclass{article}

\usepackage{microtype}
\usepackage{graphicx}
\usepackage{subfigure}
\usepackage{booktabs} 

\usepackage{hyperref}

\usepackage[accepted]{icml2025}

\usepackage{amsmath}
\usepackage{amssymb}
\usepackage{mathtools}
\usepackage{amsthm}

\usepackage[capitalize,noabbrev]{cleveref}

\usepackage{makecell} 
\usepackage{array}

\theoremstyle{plain}

\theoremstyle{definition}

\theoremstyle{remark}

\usepackage[textsize=tiny]{todonotes}

\icmltitlerunning{Enhancing CI Forecasting via CVPE Module}

\usepackage{multirow}

\begin{document}

\twocolumn[
\icmltitle{Enhancing Channel-Independent Time Series Forecasting \\
via Cross-Variate Patch Embedding}




\begin{icmlauthorlist}
\icmlauthor{Donghwa Shin}{org,uva}
\icmlauthor{Edwin Zhang}{org,comp}
\end{icmlauthorlist}

\icmlaffiliation{org}{Humanity Unleashed, Austin, TX, USA}
\icmlaffiliation{uva}{Computer Science, University of Virginia, Charlottesville, VA, USA}
\icmlaffiliation{comp}{OpenAI, USA}

\icmlcorrespondingauthor{Donghwa Shin}{gcs3ja@virginia.edu}

\icmlkeywords{Deep Learning, Time Series Forecasting, Transformer, Channel Independence (CI), Channel Dependence (CD), Patch Embedding}

\vskip 0.3in
]



\printAffiliationsAndNotice{}  

\begin{abstract}

Transformers have recently gained popularity in time series forecasting due to their ability to capture long-term dependencies. However, many existing models focus only on capturing temporal dependencies while omitting intricate relationships between variables. Recent models have tried tackling this by explicitly modeling both cross-time and cross-variate dependencies through a sequential or unified attention mechanism, but they are entirely channel dependent (CD) across all layers, making them potentially susceptible to overfitting. To address this, we propose Cross-Variate Patch Embeddings (CVPE), a lightweight CD module that injects cross-variate context into channel-independent (CI) models by simply modifying the patch embedding process. We achieve this by adding a learnable positional encoding and a lightweight router-attention block to the vanilla patch embedding layer. We then integrate CVPE into Time-LLM, a multimodal CI forecasting model, to demonstrate its effectiveness in capturing cross-variate dependencies and enhance the CI model's performance. Extensive experimental results on seven real-world datasets show that our enhanced Time-LLM outperforms the original baseline model simply by incorporating the CVPE module, with no other changes. The code is made publicly available at \url{https://github.com/Humanity-Unleashed/TimeLLM-CVPE}. 

\end{abstract}

\section{Introduction}
\label{introduction}

Time series forecasting is a critical domain in many fields, with applications in energy consumption \cite{deb2017review}, healthcare \cite{avinash2025time}, weather forecasting \cite{us_senate_2013_forecasting}, and finance \cite{gajamannage2022realtimeforecastingtimeseries}. As a result, advanced machine learning models have been employed to accurately predict future values within each domain. 

Transformers, in particular, have gained a recent surge in popularity due to their success in other sequential tasks, such as natural language processing \cite{vaswani2023attentionneed, devlin2019bertpretrainingdeepbidirectional, brown2020languagemodelsfewshotlearners, raffel2023exploringlimitstransferlearning} and computer vision \cite{dosovitskiy2021imageworth16x16words, carion2020endtoendobjectdetectiontransformers, liu2021swintransformerhierarchicalvision, he2021maskedautoencodersscalablevision}. Early time series transformers attempted to perform multivariate time series forecasting by first embedding temporal information using multivariate vectors for each time step and then applying a specialized attention mechanism \cite{zhou2021informerefficienttransformerlong, wu2022autoformerdecompositiontransformersautocorrelation, liu2022pyraformer, zhou2022fedformerfrequencyenhanceddecomposed}. However, these models were often outperformed even by simple linear models due to their vulnerability to distribution drifts \cite{zeng2022transformerseffectivetimeseries, han2023capacityrobustnesstradeoffrevisiting}, raising caution when simultaneously modeling cross-time and cross-variate dependencies. 

To address this issue, Channel Independent (CI) models have been introduced, modeling only temporal dependencies across each channel by treating a multivariate time series as a set of univariate time series forecasting tasks. For example, PatchTST transforms each univariate series into patch embeddings and independently inputs them into a vanilla transformer encoder \cite{nie2023timeseriesworth64}. Similarly, Time-LLM follows the same embedding process, but replaces the vanilla encoder with a frozen pre-trained large language model (LLM), reprogramming the patches to be compatible with the LLM \cite{jin2024timellmtimeseriesforecasting}. Despite ignoring cross-variate dependencies, CI models have shown superiority over many Channel Dependent (CD) models in both complexity and robustness \cite{han2023capacityrobustnesstradeoffrevisiting}. However, CI models cannot capture inter-variate relationships by design, implying a lower model capacity compared to CD models that can capture such dependencies. 

To mitigate the limitations of fully channel-dependent and channel-independent models, we propose a novel patch embedding strategy that incorporates cross-variate information into existing CI models using patch embeddings. Specifically, we enhance the existing patch embedding step by strategically adding a simple positional encoding and a router-attention mechanism inspired by Crossformer~\cite{zhang2023crossformer}, thus retaining the core CI backbone while improving performance. Our contributions are as follows:
\begin{itemize}
    \item We integrate a novel patch embedding step into Time-LLM to capture cross-variate dependencies, showing improved performance over the baseline with minimal changes.
    \item We demonstrate that channel-independent models can be augmented to model inter-variable interactions without requiring extensive redesign or resorting to fully CD approaches.
\end{itemize}

\section{Related Work}
\label{related work}

Due to the inherent limitations of CI models, several efforts have attempted to explicitly capture cross-variate dependencies within time series transformers \cite{liu2024itransformerinvertedtransformerseffective, zhang2023crossformer, xue2024cardchannelalignedrobust, liu2024unitsteffectivelymodelinginterseries}. iTransformer~\cite{liu2024itransformerinvertedtransformerseffective} embeds the entire univariate time series as a token and uses variate-wise attention. Crossformer~\cite{zhang2023crossformer} and CARD~\cite{xue2024cardchannelalignedrobust} both use a sequential two-stage attention mechanism to separately model cross-time and cross-variate dependencies in two steps. UniTST~\cite{liu2024unitsteffectivelymodelinginterseries} flattens all patches into one unified sequence and applies a single attention to jointly learn temporal and channel interactions. Although promising, these methodologies are fundamentally channel dependent across all layers, making them vulnerable to overfitting noise \cite{qiu2025comprehensivesurveydeeplearning}. In contrast, our approach introduces a lightweight cross-variate patch embedding (CVPE) module into a CI framework, enabling the model to learn inter-variate relationships while preserving the robustness of the CI approach. 

\section{Methodology}
\label{methodology}

In this section, we describe our CVPE module and detail its integration into a CI model to incorporate cross-variate dependencies. We adopt Time-LLM as our baseline CI model because of its modular design and state-of-the-art performance in multimodal time series forecasting. A brief overview of Time-LLM is provided in Section~\ref{sec:timellm-overview}, and a comprehensive description is provided in Appendix~\ref{appendix:timellm-detailed-overview}. 

\subsection{Brief Overview of Time-LLM}
\label{sec:timellm-overview}
Time-LLM \cite{jin2024timellmtimeseriesforecasting} is a multimodal forecasting model that "reprograms" time series data into a format compatible with a frozen LLM. First, each input channel $\mathbf{X}^{(i)}$ undergoes reversible instance normalization and is divided into $P$ time series patches of length $L_P$, creating a patched series $\mathbf{X}_P^{(i)} \in \mathbb{R}^{P \times L_P}$. These patches are then linearly projected to a dimension $d_m$, producing patch embeddings $\hat{\mathbf{X}}_P^{(i)} \in \mathbb{R}^{P \times d_m}$. Next, cross-attention layers convert these embeddings into LLM-compatible word embeddings, which are passed through the frozen LLM. Finally, the output of the LLM is flattened and projected to derive the final forecasts $\hat{\mathbf{Y}}^{(i)}$. 

Time-LLM is inherently channel independent at every layer, from input patching through LLM output flattening. Therefore, our approach carefully targets only the patch embedding step, injecting cross-variate dependencies while preserving the CI nature of all subsequent components. 

\begin{figure}[htbp]
  \centering
  \includegraphics[width=0.5\textwidth]{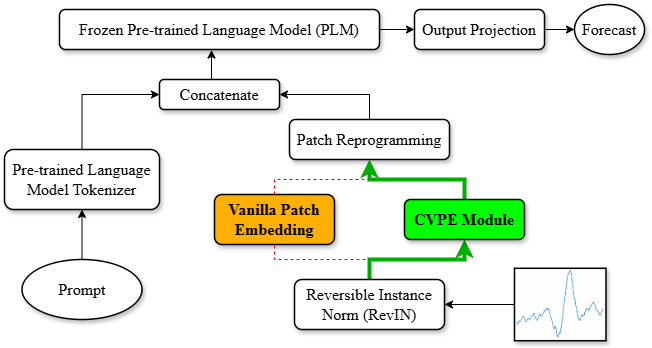}
  \caption{Model framework of Time-LLM with CVPE. Red dotted line represents original Time-LLM pipeline with the vanilla patch embedding; Green represents modified Time-LLM with CVPE module. As shown in the figure, no other modification is made to Time-LLM.}
  \label{fig:time-llm-architecture}
\end{figure}

\subsection{Cross-Variate Patch Embedding Overview}
Given a multivariate input time series with $N$ variates $\mathbf{X} \in \mathbb{R}^{N \times T}$, we first divide each channel $\mathbf{X}^{(i)}$ into $P$ patches of length $L_P$. The patches are then linearly projected onto patch embeddings with dimension $d_m$, yielding $\hat{\mathbf{X}}_P^{(i)} \in \mathbb{R}^{P \times d_m}$ that match Time-LLM's implementation. Rather than directly proceeding to the reprogramming step, we aggregate these embeddings into $\hat{\mathbf{X}}_P \in \mathbb{R}^{N \times P \times d_m}$ and apply our augmentation to introduce cross-variate dependencies at the patch level. 

\begin{figure}[htbp]
  \centering
  \includegraphics[width=0.3\textwidth]{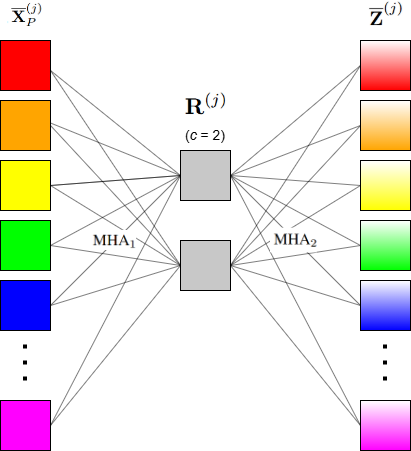}
  \caption{Visualization of router-attention mechanism, inspired by \citet{zhang2023crossformer}. Given position-aware patch embedding $\overline{\mathbf{X}}_P$, a small set of $c$ routers aggregate information from all variates, then redistribute it back to create enhanced embeddings with cross-variate information.}
  \label{fig:router-attention}
\end{figure}

We achieve this by first adding a learnable position encoding $\mathbf{W}_P \in \mathbb{R}^{P \times d_m}$ to $\hat{\mathbf{X}}_P$, embedding the positional information of each patch relative to all others across both temporal and variate dimensions. The resulting embeddings $\overline{\mathbf{X}}_P \in \mathbb{R}^{N \times P \times d_m}$ are then processed by a router-attention mechanism inspired by \citet{zhang2023crossformer}. Here, we introduce a small set of learnable router vectors $\mathbf{R} \in \mathbb{R}^{N \times c \times d_m}$ for each time step $j$, where $c$ is a constant. First, we use $\mathbf{R}$ as the query and $\overline{\mathbf{X}}_P$ as key and value in a multi-head attention (MHA) operation to produce an aggregated representation $\mathbf{A} \in \mathbb{R}^{N \times c \times d_m}$. Note that both $\mathbf{R}$ and $\overline{\mathbf{X}}_P$ are \textit{directly} used as inputs to the attention layer rather than learning separate $Q, K, V$ weight matrices. We then distribute $\mathbf{A}$ back to the patches through a second MHA, using $\overline{\mathbf{X}}_P$ as query and $\mathbf{A}$ as key and value to create cross-variate-aware patch embeddings. We outline the entire process as follows:
\begin{equation}
\begin{aligned}
\mathbf{A}^{(j)} &= \text{MHA}_1(\mathbf{R}^{(j)}, \overline{\mathbf{X}}_P^{(j)}, \overline{\mathbf{X}}_P^{(j)}) \\
\overline{\mathbf{Z}}^{(j)} &= \text{MHA}_2(\overline{\mathbf{X}}_P^{(j)}, \mathbf{A}^{(j)}, \mathbf{A}^{(j)}) \\
\hat{\mathbf{Z}} &= \text{LayerNorm}(\overline{\mathbf{X}}_P + \overline{\mathbf{Z}}) \\
\mathbf{Z} &= \text{LayerNorm}(\hat{\mathbf{Z}} + \text{MLP}(\hat{\mathbf{Z}}))
\end{aligned}
\end{equation}

where $\hat{\mathbf{Z}}$ denotes the skip connection output of the MHA operations and $\mathbf{Z}$ represents the final patch embedding after both the skip connection and MLP. Figure~\ref{fig:router-attention} provides a visual representation of this process. 

The overall complexity of the router-attention mechanism is $O(NP)$ \cite{zhang2023crossformer}, which aligns with the lightweight design of our CVPE module. 

\paragraph{Rationale}
By incorporating cross-variate information into individual time series patches, we ensure that such dependencies are retained even after the patched series is split into $N$ different series for subsequent reprogramming and LLM layers, allowing variable-wise information to propagate through the rest of the model. As a result, the output of our modified Time-LLM captures both temporal and channel dependencies present in the original input series. 

\section{Experiments}
\label{experiments}

To demonstrate the effectiveness of our patch embedding process, we conduct extensive experiments comparing CVPE-enhanced Time-LLM against the original model. We use a unified evaluation pipeline to ensure fair comparison and follow the experimental configurations provided by \citet{jin2024timellmtimeseriesforecasting}, with minor adjustments due to computational constraints. 

\subsection{Long-Term Forecasting Results}
\paragraph{Datasets}
We evaluate on ETTh1, ETTh2, ETTm1, ETTm2, Weather, Traffic, and Electricity (ECL), which have been widely adopted to benchmark long-term forecasting models \cite{wu2023timesnettemporal2dvariationmodeling} and were used to evaluate Time-LLM \cite{jin2024timellmtimeseriesforecasting}. Due to GPU memory constraints on our machines, we select the ten most correlated features from the Traffic and ECL datasets based on the Pearson correlation coefficient:
\begin{equation}
\begin{aligned}
r &= \frac{\sum (x_i - \overline{x})(y_i - \overline{y})}{\sqrt{\sum (x_i - \overline{x})^2 \sum (x_i - \overline{x})^2}} 
\end{aligned}
\end{equation}
We denote these modified datasets as \textit{Traffic (Modified)} and \textit{ECL (Modified)}, respectively. 

\paragraph{Setup}
We follow the setup used by \citet{jin2024timellmtimeseriesforecasting}, with a few key changes to address GPU memory constraints. Implementation details and hyperparameters for each dataset are provided in Appendix~\ref{appendix:experimental-details}. Unlike the original paper, we set the context window $T$ to 256 (instead of 512 in the original) to accommodate our local GPU memory limits. Similarly, we replace Llama-7B~\cite{touvron2023llamaopenefficientfoundation} with GPT-2~\cite{radford2019language} as our default LLM backbone to reduce the GPU memory footprint. The prediction horizon $H \in \{96, 192, 336, 720\}$ remains identical to \citet{jin2024timellmtimeseriesforecasting}, and we use mean squared error (MSE) and mean absolute error (MAE) as our evaluation metrics. 

\paragraph{Results}
Our brief results are provided in Table~\ref{tab:results}, where our version of Time-LLM with Cross-Variate Patch Embeddings (CVPE) outperforms the original baseline on key datasets with rich cross-variate dependencies.

\begin{table}[ht]
\label{tab:results}
\centering
\caption{Long-term forecasting results. All results are average across four different forecasting horizons: $H \in \{96,192, 336,720\}$. A lower value indicates better performance. $\bold{Bold}$: the best. Our full results for individual horizons are provided in Appendix~\ref{appendix:full-results}.}
\vspace{1em}  
\begin{tabular}{c|cc|cc}
\hline
Methods & \multicolumn{2}{c|}{\makecell{TIME-LLM w/ \\ CVPE (\textbf{Ours})}} & \multicolumn{2}{c}{\makecell{TIME-LLM \\ (Original)}} \\
\hline
Metric & MSE & MAE & MSE & MAE \\
\hline
\textit{ETTh1}   & \textbf{0.445} & \textbf{0.444} & 0.453 & 0.449 \\
\textit{ETTh2}   & 0.385 & 0.419 & \textbf{0.366} & \textbf{0.399} \\
\textit{ETTm1}   & 0.390 & 0.400 & \textbf{0.382} & \textbf{0.399} \\
\textit{ETTm2}   & 0.289 & 0.339 & \textbf{0.275} & \textbf{0.327} \\
\textit{Weather} & \textbf{0.228} & \textbf{0.267} & 0.239 & 0.273 \\
\textit{ECL (Modified)}     & \textbf{0.191} & \textbf{0.309} & 0.192 & 0.310 \\
\textit{Traffic (Modified)} & \textbf{0.126} & \textbf{0.229} & 0.135 & 0.245 \\
\hline
\end{tabular}
\end{table}

We note that our version of Time-LLM displays up to a $4.6$\% average performance gain on the Weather dataset and a $6.7$\% gain on the Modified Traffic dataset. Additionally, our version showed virtually no performance loss in the ETTh1, ETTh2, and Modified ECL datasets compared to the original Time-LLM. These results demonstrate that CI models can indeed benefit from strategically introducing cross-variate dependencies. With minimal architectural changes, our CVPE-enhanced Time-LLM delivers high performance gains on datasets with strong inter-variate correlations, such as Weather and Modified Traffic. In contrast, as expected, datasets with weaker correlations (ETTh1, ETTm1, and Modified ECL) exhibit minimal to no improvement over the original baseline. 

Despite these successful results, we note that our version underperforms on the ETTh2 and ETTm2 datasets where we saw up to a $5.2$\% performance loss. This suggests that CVPE is not entirely immune to overfitting when cross-variate dependencies are weak or irrelevant to temporal patterns. In such cases, cross-channel information may conflict with the dominant temporal interactions, negatively affecting performance.

\section{Conclusion and Future Work}
\label{conclusion}

Our study demonstrates the potential of enhancing channel-independent models through a novel cross-variate patch embedding process. We show that embedding cross-variate dependencies directly into patch embeddings can provide improved performance over the baseline on datasets with rich correlations. Despite the promising results, we observe a loss in performance on the ETTh2 and ETTm2 datasets, likely due to overfitting noise from weakly correlated features. This highlights the sensitivity of modeling cross-variate information even in lightweight modules like ours and suggests how channel mixing should be done with caution. Additionally, GPU memory constraints prevented experiments with larger datasets (e.g., Traffic and ECL) and LLMs like Llama-7B, which might have higher capacities to learn from the enriched embeddings. Future research should address these by assessing CVPE with larger backbones and evaluating performance on the full ECL and Traffic datasets. Furthermore, future work should explore alternative strategies to our current router attention block to better mitigate overfitting, such as attention mechanisms that attend to only a subset of channels (channel partiality) rather than all.

\bibliography{source}

\begin{thebibliography}{29}
\providecommand{\natexlab}[1]{#1}
\providecommand{\url}[1]{\texttt{#1}}
\expandafter\ifx\csname urlstyle\endcsname\relax
  \providecommand{\doi}[1]{doi: #1}\else
  \providecommand{\doi}{doi: \begingroup \urlstyle{rm}\Url}\fi

\bibitem[Avinash et~al.(2025)Avinash, Pachori, Sharma, and Mishra]{avinash2025time}
Avinash, G., Pachori, H., Sharma, A., and Mishra, S.
\newblock Time series forecasting of bed occupancy in mental health facilities in india using machine learning.
\newblock \emph{Scientific Reports}, 15\penalty0 (2686), 2025.
\newblock \doi{10.1038/s41598-025-86418-9}.
\newblock URL \url{https://www.nature.com/articles/s41598-025-86418-9}.

\bibitem[Brown et~al.(2020)Brown, Mann, Ryder, Subbiah, Kaplan, Dhariwal, Neelakantan, Shyam, Sastry, Askell, Agarwal, Herbert-Voss, Krueger, Henighan, Child, Ramesh, Ziegler, Wu, Winter, Hesse, Chen, Sigler, Litwin, Gray, Chess, Clark, Berner, McCandlish, Radford, Sutskever, and Amodei]{brown2020languagemodelsfewshotlearners}
Brown, T.~B., Mann, B., Ryder, N., Subbiah, M., Kaplan, J., Dhariwal, P., Neelakantan, A., Shyam, P., Sastry, G., Askell, A., Agarwal, S., Herbert-Voss, A., Krueger, G., Henighan, T., Child, R., Ramesh, A., Ziegler, D.~M., Wu, J., Winter, C., Hesse, C., Chen, M., Sigler, E., Litwin, M., Gray, S., Chess, B., Clark, J., Berner, C., McCandlish, S., Radford, A., Sutskever, I., and Amodei, D.
\newblock Language models are few-shot learners, 2020.
\newblock URL \url{https://arxiv.org/abs/2005.14165}.

\bibitem[Carion et~al.(2020)Carion, Massa, Synnaeve, Usunier, Kirillov, and Zagoruyko]{carion2020endtoendobjectdetectiontransformers}
Carion, N., Massa, F., Synnaeve, G., Usunier, N., Kirillov, A., and Zagoruyko, S.
\newblock End-to-end object detection with transformers, 2020.
\newblock URL \url{https://arxiv.org/abs/2005.12872}.

\bibitem[Deb et~al.(2017)Deb, Zhang, Yang, Lee, and Shah]{deb2017review}
Deb, C., Zhang, F., Yang, J., Lee, S.~E., and Shah, K.~W.
\newblock A review on time series forecasting techniques for building energy consumption.
\newblock \emph{Renewable and Sustainable Energy Reviews}, 74:\penalty0 902--924, 2017.
\newblock \doi{10.1016/j.rser.2017.02.085}.
\newblock URL \url{https://daneshyari.com/article/preview/5483166.pdf}.

\bibitem[Devlin et~al.(2019)Devlin, Chang, Lee, and Toutanova]{devlin2019bertpretrainingdeepbidirectional}
Devlin, J., Chang, M.-W., Lee, K., and Toutanova, K.
\newblock Bert: Pre-training of deep bidirectional transformers for language understanding, 2019.
\newblock URL \url{https://arxiv.org/abs/1810.04805}.

\bibitem[Dosovitskiy et~al.(2021)Dosovitskiy, Beyer, Kolesnikov, Weissenborn, Zhai, Unterthiner, Dehghani, Minderer, Heigold, Gelly, Uszkoreit, and Houlsby]{dosovitskiy2021imageworth16x16words}
Dosovitskiy, A., Beyer, L., Kolesnikov, A., Weissenborn, D., Zhai, X., Unterthiner, T., Dehghani, M., Minderer, M., Heigold, G., Gelly, S., Uszkoreit, J., and Houlsby, N.
\newblock An image is worth 16x16 words: Transformers for image recognition at scale, 2021.
\newblock URL \url{https://arxiv.org/abs/2010.11929}.

\bibitem[Gajamannage \& Park(2022)Gajamannage and Park]{gajamannage2022realtimeforecastingtimeseries}
Gajamannage, K. and Park, Y.
\newblock Real-time forecasting of time series in financial markets using sequentially trained many-to-one lstms, 2022.
\newblock URL \url{https://arxiv.org/abs/2205.04678}.

\bibitem[Han et~al.(2023)Han, Ye, and Zhan]{han2023capacityrobustnesstradeoffrevisiting}
Han, L., Ye, H.-J., and Zhan, D.-C.
\newblock The capacity and robustness trade-off: Revisiting the channel independent strategy for multivariate time series forecasting, 2023.
\newblock URL \url{https://arxiv.org/abs/2304.05206}.

\bibitem[He et~al.(2021)He, Chen, Xie, Li, Dollár, and Girshick]{he2021maskedautoencodersscalablevision}
He, K., Chen, X., Xie, S., Li, Y., Dollár, P., and Girshick, R.
\newblock Masked autoencoders are scalable vision learners, 2021.
\newblock URL \url{https://arxiv.org/abs/2111.06377}.

\bibitem[Jin et~al.(2024)Jin, Wang, Ma, Chu, Zhang, Shi, Chen, Liang, Li, Pan, and Wen]{jin2024timellmtimeseriesforecasting}
Jin, M., Wang, S., Ma, L., Chu, Z., Zhang, J.~Y., Shi, X., Chen, P.-Y., Liang, Y., Li, Y.-F., Pan, S., and Wen, Q.
\newblock Time-llm: Time series forecasting by reprogramming large language models, 2024.
\newblock URL \url{https://arxiv.org/abs/2310.01728}.

\bibitem[Kingma \& Ba(2017)Kingma and Ba]{kingma2017adammethodstochasticoptimization}
Kingma, D.~P. and Ba, J.
\newblock Adam: A method for stochastic optimization, 2017.
\newblock URL \url{https://arxiv.org/abs/1412.6980}.

\bibitem[Liu et~al.(2024{\natexlab{a}})Liu, Liu, Woo, Wang, Hooi, Xiong, and Sahoo]{liu2024unitsteffectivelymodelinginterseries}
Liu, J., Liu, C., Woo, G., Wang, Y., Hooi, B., Xiong, C., and Sahoo, D.
\newblock Unitst: Effectively modeling inter-series and intra-series dependencies for multivariate time series forecasting, 2024{\natexlab{a}}.
\newblock URL \url{https://arxiv.org/abs/2406.04975}.

\bibitem[Liu et~al.(2022)Liu, Yu, Liao, Li, Lin, Liu, and Dustdar]{liu2022pyraformer}
Liu, S., Yu, H., Liao, C., Li, J., Lin, W., Liu, A.~X., and Dustdar, S.
\newblock Pyraformer: Low-complexity pyramidal attention for long-range time series modeling and forecasting.
\newblock In \emph{International Conference on Learning Representations}, 2022.
\newblock URL \url{https://openreview.net/forum?id=0EXmFzUn5I}.

\bibitem[Liu et~al.(2024{\natexlab{b}})Liu, Hu, Zhang, Wu, Wang, Ma, and Long]{liu2024itransformerinvertedtransformerseffective}
Liu, Y., Hu, T., Zhang, H., Wu, H., Wang, S., Ma, L., and Long, M.
\newblock itransformer: Inverted transformers are effective for time series forecasting, 2024{\natexlab{b}}.
\newblock URL \url{https://arxiv.org/abs/2310.06625}.

\bibitem[Liu et~al.(2021)Liu, Lin, Cao, Hu, Wei, Zhang, Lin, and Guo]{liu2021swintransformerhierarchicalvision}
Liu, Z., Lin, Y., Cao, Y., Hu, H., Wei, Y., Zhang, Z., Lin, S., and Guo, B.
\newblock Swin transformer: Hierarchical vision transformer using shifted windows, 2021.
\newblock URL \url{https://arxiv.org/abs/2103.14030}.

\bibitem[Nie et~al.(2023)Nie, Nguyen, Sinthong, and Kalagnanam]{nie2023timeseriesworth64}
Nie, Y., Nguyen, N.~H., Sinthong, P., and Kalagnanam, J.
\newblock A time series is worth 64 words: Long-term forecasting with transformers, 2023.
\newblock URL \url{https://arxiv.org/abs/2211.14730}.

\bibitem[Qiu et~al.(2025)Qiu, Cheng, Wu, Hu, Guo, and Yang]{qiu2025comprehensivesurveydeeplearning}
Qiu, X., Cheng, H., Wu, X., Hu, J., Guo, C., and Yang, B.
\newblock A comprehensive survey of deep learning for multivariate time series forecasting: A channel strategy perspective, 2025.
\newblock URL \url{https://arxiv.org/abs/2502.10721}.

\bibitem[Radford et~al.(2019)Radford, Wu, Child, Luan, Amodei, and Sutskever]{radford2019language}
Radford, A., Wu, J., Child, R., Luan, D., Amodei, D., and Sutskever, I.
\newblock Language models are unsupervised multitask learners.
\newblock \emph{OpenAI}, 2019.
\newblock URL \url{https://cdn.openai.com/better-language-models/language_models_are_unsupervised_multitask_learners.pdf}.

\bibitem[Raffel et~al.(2023)Raffel, Shazeer, Roberts, Lee, Narang, Matena, Zhou, Li, and Liu]{raffel2023exploringlimitstransferlearning}
Raffel, C., Shazeer, N., Roberts, A., Lee, K., Narang, S., Matena, M., Zhou, Y., Li, W., and Liu, P.~J.
\newblock Exploring the limits of transfer learning with a unified text-to-text transformer, 2023.
\newblock URL \url{https://arxiv.org/abs/1910.10683}.

\bibitem[Touvron et~al.(2023)Touvron, Lavril, Izacard, Martinet, Lachaux, Lacroix, Rozière, Goyal, Hambro, Azhar, Rodriguez, Joulin, Grave, and Lample]{touvron2023llamaopenefficientfoundation}
Touvron, H., Lavril, T., Izacard, G., Martinet, X., Lachaux, M.-A., Lacroix, T., Rozière, B., Goyal, N., Hambro, E., Azhar, F., Rodriguez, A., Joulin, A., Grave, E., and Lample, G.
\newblock Llama: Open and efficient foundation language models, 2023.
\newblock URL \url{https://arxiv.org/abs/2302.13971}.

\bibitem[{United States Senate Committee on Commerce, Science, and Transportation}(2013)]{us_senate_2013_forecasting}
{United States Senate Committee on Commerce, Science, and Transportation}.
\newblock Forecasting success: Achieving u.s. weather readiness for the long term.
\newblock Hearing before the Subcommittee on Oceans, Atmosphere, Fisheries, and Coast Guard, December 2013.
\newblock URL \url{https://www.govinfo.gov/content/pkg/CHRG-113shrg93659/pdf/CHRG-113shrg93659.pdf}.
\newblock S. Hrg. 113–594.

\bibitem[Vaswani et~al.(2023)Vaswani, Shazeer, Parmar, Uszkoreit, Jones, Gomez, Kaiser, and Polosukhin]{vaswani2023attentionneed}
Vaswani, A., Shazeer, N., Parmar, N., Uszkoreit, J., Jones, L., Gomez, A.~N., Kaiser, L., and Polosukhin, I.
\newblock Attention is all you need, 2023.
\newblock URL \url{https://arxiv.org/abs/1706.03762}.

\bibitem[Wu et~al.(2022)Wu, Xu, Wang, and Long]{wu2022autoformerdecompositiontransformersautocorrelation}
Wu, H., Xu, J., Wang, J., and Long, M.
\newblock Autoformer: Decomposition transformers with auto-correlation for long-term series forecasting, 2022.
\newblock URL \url{https://arxiv.org/abs/2106.13008}.

\bibitem[Wu et~al.(2023)Wu, Hu, Liu, Zhou, Wang, and Long]{wu2023timesnettemporal2dvariationmodeling}
Wu, H., Hu, T., Liu, Y., Zhou, H., Wang, J., and Long, M.
\newblock Timesnet: Temporal 2d-variation modeling for general time series analysis, 2023.
\newblock URL \url{https://arxiv.org/abs/2210.02186}.

\bibitem[Xue et~al.(2024)Xue, Zhou, Wen, Gao, Ding, and Jin]{xue2024cardchannelalignedrobust}
Xue, W., Zhou, T., Wen, Q., Gao, J., Ding, B., and Jin, R.
\newblock Card: Channel aligned robust blend transformer for time series forecasting, 2024.
\newblock URL \url{https://arxiv.org/abs/2305.12095}.

\bibitem[Zeng et~al.(2022)Zeng, Chen, Zhang, and Xu]{zeng2022transformerseffectivetimeseries}
Zeng, A., Chen, M., Zhang, L., and Xu, Q.
\newblock Are transformers effective for time series forecasting?, 2022.
\newblock URL \url{https://arxiv.org/abs/2205.13504}.

\bibitem[Zhang \& Yan(2023)Zhang and Yan]{zhang2023crossformer}
Zhang, Y. and Yan, J.
\newblock Crossformer: Transformer utilizing cross-dimension dependency for multivariate time series forecasting.
\newblock In \emph{The Eleventh International Conference on Learning Representations}, 2023.
\newblock URL \url{https://openreview.net/forum?id=vSVLM2j9eie}.

\bibitem[Zhou et~al.(2021)Zhou, Zhang, Peng, Zhang, Li, Xiong, and Zhang]{zhou2021informerefficienttransformerlong}
Zhou, H., Zhang, S., Peng, J., Zhang, S., Li, J., Xiong, H., and Zhang, W.
\newblock Informer: Beyond efficient transformer for long sequence time-series forecasting, 2021.
\newblock URL \url{https://arxiv.org/abs/2012.07436}.

\bibitem[Zhou et~al.(2022)Zhou, Ma, Wen, Wang, Sun, and Jin]{zhou2022fedformerfrequencyenhanceddecomposed}
Zhou, T., Ma, Z., Wen, Q., Wang, X., Sun, L., and Jin, R.
\newblock Fedformer: Frequency enhanced decomposed transformer for long-term series forecasting, 2022.
\newblock URL \url{https://arxiv.org/abs/2201.12740}.

\end{thebibliography}
\bibliographystyle{icml2025}

\newpage
\appendix
\onecolumn
\section{Detailed Overview of Time-LLM Architecture}
\label{appendix:timellm-detailed-overview}

\paragraph{Input Embedding}
Given an input channel $\mathbf{X}^{(i)}$, Time-LLM~\cite{jin2024timellmtimeseriesforecasting} starts by applying a reversible instance normalization (RevIN) on the given series, normalizing it with zero mean and unit standard deviation. The model then divides $\mathbf{X}^{(i)}$ into $P$ patches of length $L_P$, where $P = \lfloor \frac{(T-L_P)}{S} \rfloor$ where $S$ is the horizontal sliding stride. The resulting patched series $\mathbf{X}_P^{(i)} \in \mathbb{R}^{P \times L_P}$ is then linearly projected onto dimension $d_m$, creating patch embeddings $\hat{\mathbf{X}}_P^{(i)} \in \mathbb{R}^{P \times d_m}$.

\paragraph{Patch Reprogramming}
During reprogramming, Time-LLM prepares pre-trained word embeddings $E \in \mathbb{R}^{V \times D}$, where $V$ is the vocabulary size. The model applies linear probing to these embeddings, generating text prototypes $E' \in \mathbb{R}^{V' \times D}$, where $V' \ll V$. Then, a multi-head cross-attention layer is employed with $\hat{\mathbf{X}}_P$ as query and $E'$ as key and value, effectively reprogramming the input series into text prototypes that the LLM can easily interpret. Lastly, the resulting reprogrammed series is linearly projected to $D$ to ensure compatibility with the LLM backbone, producing $\mathbf{O}^{(i)} \in \mathbb{R}^{P \times D}$. 

\paragraph{LLM Backbone and Output Projection}
Once reprogramming is complete, the resulting series is processed through the LLM backbone. The output of the LLM is post-processed by removing prefixal parts, flattening the output series, linearly projecting it, and applying denormalization to derive final forecasts $\mathbf{Y}^{(i)}$. 

\section{Experimental Details}
\label{appendix:experimental-details}

\subsection{Dataset Details}
Dataset statistics are summarized in Table~\ref{tab:dataset-details}. We evaluate long-term forecasting performance on five established benchmarks, including four ETT datasets~\cite{zhou2021informerefficienttransformerlong} (i.e., ETTh1, ETTh2, ETTm1, and ETTm2) and Weather~\cite{wu2023timesnettemporal2dvariationmodeling}. We also use a modified version of the Electricity and Traffic datasets~\cite{wu2023timesnettemporal2dvariationmodeling}, extracting only the top ten correlated features. Specifically, we choose variables $\{101, 173, 267, 280, 440, 441, 721, 842, 857, OT\}$ from the Traffic dataset and label it as \textit{Traffic (Modified)}; similarly, we extract $\{112, 117, 119, 123, 152, 158, 173, 278, 313, OT\}$ from the Electricity dataset and label it as \textit{ECL (Modified)}.

\begin{table}[ht]
\label{tab:dataset-details}
\centering
\caption{Dataset statistics are from \cite{wu2023timesnettemporal2dvariationmodeling} and adopted by \citet{jin2024timellmtimeseriesforecasting}. Channel indicates the total number of variables in the dataset, and the dataset size is split into (training, validation, testing).}
\vspace{1em}
\begin{tabular}{c|c|c|c|c|c}
\hline
Dataset & Channel & Dataset Size & Frequency & Domain \\
\hline
ETTh1 & 7  & (8545, 2881, 2881) & 1 hour & Temperature \\
\hline
ETTh2 & 7  & (8545, 2881, 2881) & 1 hour & Temperature \\
\hline
ETTm1 & 7  & (34465, 11521, 11521) & 15 min & Temperature \\
\hline
ETTm2 & 7  & (34465, 11521, 11521) & 15 min & Temperature \\
\hline
Weather & 21  & (36792, 5271, 10540) & 10 min & Weather \\
\hline
ECL (Modified) & 10  & (18317, 2633, 5261) & 1 hour & Electricity \\
\hline
Traffic (Modified) & 10  & (12185, 1757, 3509) & 1 hour & Transportation \\
\hline
\end{tabular}
\end{table}

\subsection{Evaluation Metrics}
We utilize the mean squared error (MSE) and mean absolute error (MAE) as our primary metrics for evaluating long-term forecasting performance. The calculation of these metrics are as follows:
\[
\begin{array}{ll}
\text{MSE} = \dfrac{1}{H} \sum\limits_{h=1}^{H} (Y_h - \hat{Y}_h)^2, & 
\text{MAE} = \dfrac{1}{H} \sum\limits_{h=1}^{H} |Y_h - \hat{Y}_h|
\end{array}
\]
where $H$ denotes the prediction horizon (i.e., number of data points). $Y_h$ and $\hat{Y_h}$ represent the $h$-th ground truth and prediction where $h \in \{1, ..., H\}$.

\subsection{Model Configurations}
We compile an overview of our model configurations in Table~\ref{tab:model-configurations}. We mostly follow the configurations used by \citet{jin2024timellmtimeseriesforecasting} but with key changes to accommodate GPU memory constraints in our local compute environment. Specifically, we alter the LLM backbone from Llama-7B to GPT-2 and reduce the input length $T$ from $512$ to $256$. Hyperparameters, such as batch size and epoch count, are also adjusted to ensure consistency, while the Adam optimizer \cite{kingma2017adammethodstochasticoptimization} is used throughout all experiments to mirror \citet{jin2024timellmtimeseriesforecasting}. Heads $K$ denote the number of heads in the multi-head attention layers used in CVPE's router attention mechanism, as well as the reprogramming layer. 

\begin{table}[ht]
\centering
\caption{Experimental configurations for Time-LLM (both CVPE-enhanced and original).}
\vspace{1em}
\label{tab:model-configurations}
\begin{tabular}{c|c|c|c|c|c|c|c|c}
\hline
Dataset & Text Prototype $V'$ & $T$ & $d_m$ & Heads $K$ & Initial Learning Rate & Loss & Batch & Epochs \\
\hline
ETTh1 & 1000 & 256 & 32 & 8 & $10^{-2}$ & MSE & 8 & 100 \\
\hline
ETTh2 & 1000 & 256 & 32 & 8 & $10^{-2}$ & MSE & 8 & 100 \\
\hline
ETTm1 & 1000 & 256 & 32 & 8 & $10^{-2}$ & MSE & 8 & 100 \\
\hline
ETTm2 & 1000 & 256 & 32 & 8 & $10^{-2}$ & MSE & 8 & 100 \\
\hline
Weather & 1000 & 256 & 32 & 8 & $10^{-2}$ & MSE & 8 & 10 \\
\hline
Electricity & 1000 & 256 & 32 & 8 & $10^{-2}$ & MSE & 8 & 100 \\
\hline
Traffic & 1000 & 256 & 32 & 8 & $10^{-2}$ & MSE & 8 & 100 \\
\hline
\end{tabular}
\end{table}

\clearpage  

\section{Long-Term Forecasting Full Results}
\label{appendix:full-results}

By simply replacing vanilla patch embeddings with the CVPE module, our modified Time-LLM variant outperforms the original on datasets exhibiting high variate-wise correlation. This result highlights the value of injecting cross-variate information directly into patch embeddings to enhance the capacity and performance of channel-independent models. In Table~\ref{tab:full-results}, we benchmark our CVPE-enhanced variant against the original baseline on four forecast horizons. Notably, our modified variant consistently produces superior results across different horizons on strongly correlated datasets (e.g., Weather and Modified Traffic), while showing no performance loss on weakly correlated datasets (e.g., ETTh1, ETTm1, and Modified ECL). Despite these promising results, the poor performance across most horizons in ETTh2 and ETTm2 suggests potential overfitting in our CVPE design. 

\begin{table}[ht]
\label{tab:full-results}
\centering
\caption{Full long-term forecasting results with horizon length $H \in \{96, 192, 336, 720\}$ for all datasets. A lower value indicates better performance. $\bold{Bold}$: the best.}
\vspace{1em}
\begin{tabular}{c|c|cc|cc}
\hline
Methods & Horizon & \multicolumn{2}{c|}{\makecell{TIME-LLM w/ \\ CVPE (\textbf{Ours})}} & \multicolumn{2}{c}{\makecell{TIME-LLM \\ (Original)}} \\
\hline
Metric & & MSE & MAE & MSE & MAE \\
\hline
\multirow{4}{*}{\textit{ETTh1}} 
  & 96  & 0.387 & 0.403 & \textbf{0.380} & \textbf{0.399} \\
  & 192 & 0.440 & 0.436 & \textbf{0.435} & \textbf{0.433} \\
  & 336 & 0.485 & 0.464 & \textbf{0.470} & \textbf{0.454} \\
  & 720 & \textbf{0.467} & \textbf{0.472} & 0.526 & 0.509 \\
\hline
\multirow{4}{*}{\textit{ETTh2}} 
  & 96  & \textbf{0.299} & \textbf{0.351} & 0.303 & 0.353 \\
  & 192 & 0.417 & 0.440 & \textbf{0.374} & \textbf{0.398} \\
  & 336 & 0.396 & 0.432 & \textbf{0.386} & \textbf{0.413} \\
  & 720 & 0.428 & 0.451 & \textbf{0.400} & \textbf{0.431} \\
\hline
\multirow{4}{*}{\textit{ETTm1}} 
  & 96  & \textbf{0.329} & \textbf{0.367} & 0.332 & 0.372 \\
  & 192 & 0.359 & \textbf{0.382} & 0.359 & 0.389 \\
  & 336 & 0.415 & 0.412 & \textbf{0.396} & \textbf{0.404} \\
  & 720 & 0.455 & 0.439 & \textbf{0.439} & \textbf{0.432} \\
\hline
\multirow{4}{*}{\textit{ETTm2}} 
  & 96  & 0.173 & \textbf{0.259} & 0.173 & 0.260 \\
  & 192 & 0.254 & 0.322 & \textbf{0.244} & \textbf{0.312} \\
  & 336 & 0.307 & 0.355 & \textbf{0.291} & \textbf{0.340} \\
  & 720 & 0.422 & 0.421 & \textbf{0.391} & \textbf{0.397} \\
\hline
\multirow{4}{*}{\textit{Weather}} 
  & 96  & \textbf{0.149} & \textbf{0.200} & 0.164 & 0.212 \\
  & 192 & \textbf{0.191} & \textbf{0.241} & 0.206 & 0.251 \\
  & 336 & \textbf{0.246} & \textbf{0.285} & 0.257 & 0.289 \\
  & 720 & \textbf{0.324} & 0.341 & 0.330 & \textbf{0.339} \\
\hline
\multirow{4}{*}{\textit{ECL (Modified)}} 
  & 96  & 0.154 & 0.276 & \textbf{0.151} & \textbf{0.271} \\
  & 192 & \textbf{0.164} & \textbf{0.282} & 0.167 & 0.286 \\
  & 336 & 0.194 & 0.312 & \textbf{0.193} & \textbf{0.311} \\
  & 720 & \textbf{0.250} & \textbf{0.367} & 0.255 & 0.372 \\
\hline
\multirow{4}{*}{\textit{Traffic (Modified)}} 
  & 96  & \textbf{0.124} & \textbf{0.231} & 0.126 & 0.237 \\
  & 192 & \textbf{0.117} & \textbf{0.217} & 0.124 & 0.233 \\
  & 336 & \textbf{0.126} & \textbf{0.229} & 0.131 & 0.239 \\
  & 720 & \textbf{0.137} & \textbf{0.240} & 0.158 & 0.270 \\
\hline
\end{tabular}
\end{table}

\end{document}